\documentclass[conference]{IEEEtran}
\IEEEoverridecommandlockouts

\usepackage{amsmath,amssymb,amsfonts,bbm}
\usepackage{algorithmic}
\usepackage{graphicx}
\usepackage{textcomp}
\usepackage{xcolor}
\usepackage{hyperref}
\usepackage[inline]{enumitem}

\usepackage{cite}

\DeclareMathOperator{\argmin}{argmin}

\makeatletter
\newcommand{\linebreakand}{%
  \end{@IEEEauthorhalign}
  \hfill\mbox{}\par
  \mbox{}\hfill\begin{@IEEEauthorhalign}
}
\makeatother

\begin{document}
\renewcommand{\baselinestretch}{1}

\title{Finding Archetypal Spaces Using Neural Networks\\
\thanks{
*,$\dagger$ Equal contribution. $\ddagger$ Corresponding author. This research was partially supported by NIH award F31HD097958 [\emph{D.B.}]; IVADO (l'institut de valorisation des donn\'{e}es) [\emph{G.W.}]; and NIH grant R01GM130847 [\emph{S.K.}].}
}

\author{\IEEEauthorblockN{David van Dijk$^{*}$}
\IEEEauthorblockA{\textit{Internal Medicine; Computer Science} \\
\textit{Yale University}\\
New Haven, CT \\
david.vandijk@yale.edu}
\and
\IEEEauthorblockN{Daniel B. Burkhardt$^{*}$}
\IEEEauthorblockA{\textit{Genetics} \\
\textit{Yale University}\\
New Haven, CT \\
daniel.burkhardt@yale.edu}
\and
\IEEEauthorblockN{Matthew Amodio}
\IEEEauthorblockA{\textit{Computer Science} \\
\textit{Yale University}\\
New Haven, CT \\
matthew.amodio@yale.edu}
\linebreakand
\IEEEauthorblockN{Alexander Tong}
\IEEEauthorblockA{\textit{Computer Science} \\
\textit{Yale University}\\
New Haven, CT \\
alexander.tong@yale.edu}
\and
\IEEEauthorblockN{Guy Wolf$^{\dagger}$}
\IEEEauthorblockA{\textit{Mathematics \& Statistics} \\
\textit{Univ.\ de Montr\'{e}al; Mila}\\
Montreal, Canada \\
guy.wolf@umontreal.ca}
\and
\IEEEauthorblockN{Smita Krishnaswamy$^{\dagger\ddagger}$}
\IEEEauthorblockA{\textit{Genetics; Computer Science} \\
\textit{Yale University}\\
New Haven, CT \\
smita.krishnaswamy@yale.edu}
}


\maketitle

\IEEEpubidadjcol

\begin{abstract}
Archetypal analysis is a data decomposition method that describes each observation in a dataset as a convex combination of "pure types" or archetypes. These archetypes represent extrema of a data space in which there is a trade-off between features, such as in biology where different combinations of traits provide optimal fitness for different environments. Existing methods for archetypal analysis work well when a linear relationship exists between the feature space and the archetypal space. However, such methods are not applicable to systems where the feature space is generated non-linearly from the combination of archetypes, such as in biological systems or image transformations. Here, we propose a reformulation of the problem such that the goal is to learn a non-linear transformation of the data into a latent archetypal space. To solve this problem, we introduce Archetypal Analysis network (AAnet), which is a deep neural network framework for learning and generating from a latent archetypal representation of data. We demonstrate state-of-the-art recovery of ground-truth archetypes in non-linear data domains, show AAnet can generate from data geometry rather than from data density, and use AAnet to identify biologically meaningful archetypes in single-cell gene expression data.

\end{abstract}


\section{Introduction}

Archetypal analysis (AA) decomposes each observation in a dataset into a convex combination of pure types or \textit{archetypes}. These archetypes represent extreme combinations of features and thus are extrema of the data space. For example, species adapted to specific environments will have unique and extremal combinations of features~\cite{shoval2012evolutionary}. Since each observation is described as a mixture of the archetypes, AA describes the dataset as varying smoothly between the identified archetypes. This interpretation has several applications for exploratory data analysis. For example, the archetypes can be characterized in the feature space to understand the extrema of a dataset. Additionally, when considering the \textit{archetypal space}, \textit{i.e.} the mixture of archetypes for each data point, AA provides a new factor space for data exploration. A point can now be characterized by its composition of specific archetypes, and distances between points can be calculated from archetypal mixtures. These applications have led to the application of AA for exploratory data analysis in a number of disciplines including astronomy \cite{chanArchetypalAnalysisGalaxy2003}, market research \cite{li2003archetypal, porzioUseArchetypesBenchmarks2008}, document analysis \cite{sethProbabilisticArchetypalAnalysis2016, canhasiWeightedHierarchicalArchetypal2016}, and genomic inference \cite{thogersenArchetypalAnalysisDiverse2013, korem2015geometry, hartInferringBiologicalTasks2015}.

Because each point is represented as a convex combination of archetypes, there is an inherent trade-off between the archetypes. This limits the number of archetypes identifiable in $\mathbb{R}^{n}$ to $n+1$. It is not possible to fit four archetypes to a rectangle in $\mathbb{R}^{2}$. This constraint well fits systems with an inherent trade-off between features, such as in genomics where typically only relative abundances of genes are considered \cite{hartInferringBiologicalTasks2015}. In this way, AA bears similarity to Latent Dirichlet Allocation (LDA), a statistical method used for topic analysis that models word occurrences in a document as occurring with some probability over a discrete number of topics with a Dirichlet prior~\cite{bleiLatentDirichletAllocation2003}. Thus, the latent features in LDA also form a space bound by a simplex. However in LDA, the topics are known \textit{a priori}, and the goal of AA is to identify the archetypes. Finally, AA implies a data model where each point varies continuously between a set of archetypes, unlike the model of clustering methods where data originates from centroids plus noise. For such cluster-like data sets, AA would need to be applied to each cluster independently.


Identifying archetypes is the primary challenge in AA. Most methods for AA identify archetypes by fitting a simplex to the data space where the vertices are linear combinations of the input data. A limitation of this approach is that if the relationships between features in the dataset are non-linear, then the extrema of the data space may not correspond to the extrema of the data geometry. Take, for example, a triangle projected onto a sphere. Although the vertices of the triangle remain the extrema of the data geometry, they may no longer conform to extrema of the data space (Fig.~\ref{fig:triangle}). In this case, linear AA methods fail to capture correct archetypes as shown in Section \ref{sec:results/triangle}. Non-linear AA methods have been proposed, such as kernel PCHA~\cite{morup2012archetypal}. However, in these methods a fixed non-linear transformation is applied to the data after which linear AA is performed. There is no guarantee that any one transformation makes all data sets well-approximated by a simplex.

To overcome these limitations, we propose a new formulation of the problem. Instead of fitting a convex hull to a fixed feature space, our goal is to identify a transformation of feature space $\mathbf{X}$ into an \textit{k}-dimensional archetypal space where \textit{k} corresponds to the number of archetypes. In the archetypal space, $\mathbf{Z}$, single activations of each dimension correspond to archetypes (\textit{i.e.} [1,0,0] for a space with 3 archetypes). The space is constrained such that each data point is represented as a convex combination of the archetypes. Because of the convexity constraint, all observations are bound by a \textit{k}-dimensional simplex. In this reformulation, the goal of AA is to learn the ideal transformation $f(\mathbf{X}) \rightarrow \mathbf{Z}$ and inverse function $f^{\prime}(\mathbf{Z}) \rightarrow  \mathbf{X}$ such that the underlying data geometry is preserved.

To achieve this, we introduce the Archetypal Analysis network (AAnet), a neural network framework for learning and generating from a latent archetypal space. AAnet uses an autoencoder with a novel regularization on the latent layer in which the encoder $E$ learns the transformation from the data space (input) to the archetypal space (bottleneck layer), and the decoder $D$ learns the transformation back to the feature space (reconstruction). Performing AA in this manner also provides powerful generative properties. Single activations of each node in the latent space represent an archetype of the data that the decoder transforms back to the feature space. It is also possible to generate new data with a specific mixture of each archetype. In contrast, the latent space of generative models such as the VAE or the sampling space of a GAN have no accessible semantic structure from which to generate data as a mixture of “pure types”. Furthermore, AAnet can sample from the data geometry independent of data density, which are limitations of VAEs and GANs.

The main contributions of this paper are:
\begin{enumerate*}
\item A reformulation of archetypal analysis with the goal of learning an optimal transformation of the data in the feature space into an archetypal space bound by a simplex;
\item A novel regularization on the latent space of an autoencoder such that nodes of the bottleneck layer are archetypes and node activations are loadings of the data onto the archetypes;
\item Demonstration of the generative properties of AAnet on unevenly sampled data with comparisons to a VAE and GAN; and
\item An extensive collection of quantitative benchmarks comparing AAnet against five state-of-the-art archetypal analysis methods.
\end{enumerate*}

The remainder of the paper provides a summary of previous work, description of the AAnet framework and implementation, quantitative comparisons of AAnet to existing AA methods on synthetic datasets, application of AAnet to a new single-cell gene expression dataset, and demonstrations of the reproducibility, robustness, and scalability of AAnet.

\section{Previous work and Background}
\label{sec:prev-approaches}

The first algorithm proposed for archetypal analysis was principal convex hull analysis (PCHA) as described by \cite{cutler1994archetypal}, which identifies a set of $p$ archetypes constrained to be linear combinations of the data such that the following is minimized:
\begin{equation}
\underset{\mathbf{W,H}}{\min}||\mathbf{X^{\prime}} - \mathbf{X^{\prime}WH}||^{2}_{F} \label{eqn:PCHA}
\end{equation}

Here, $\mathbf{X}$ is the data matrix with $n$ observations on the rows and $m$ features. $\mathbf{W}$ is an $n \times p$ matrix mapping the data to the archetypes and $\mathbf{W}$ is a $p \times n$ matrix denoting the archetypes in the feature space. Cutler and Breiman \cite{cutler1994archetypal} then propose an optimization algorithm using alternating least squares.

Subsequent advances focused on improvements to the algorithm for fitting a hull to the data. In \cite{morup2012archetypal}, it is proposed to solve the PCHA optimization via projected gradient descent. Further improvement to the optimization procedures are formed in \cite{chen2014fast}, which uses an active set strategy. More recently, envelope constraints were tightened in \cite{javadi2017non} by adding a cost for the sum of the distances of the data points from the convex envelope of the archetypes and another for the sum of the distances of archetypes from the convex envelope of the data points. 

The first work to propose AA on a transformed feature space is \cite{morup2012archetypal}. There, an algorithm is provided for AA applied to the kernel space of a dataset. In \cite{wynen2018unsupervised}, the authors perform archetypal analysis on the representation found in a hidden layer of an image classification neural network in order to define image styles. Although these methods extend AA to non-linear feature spaces, both apply a fixed transformation to the data space. By contrast, our goal is to \textit{find} an optimal non-linear transformation of the data such that the data is optimally described by a simplex. We propose to use a novel neural network regularization for this task.


\section{Methods}

First, we describe our new generalized problem formulation for finding a transformed data space for archetypal analysis, and then we describe our AAnet framework. 

\subsection{Problem setup}
\label{sec:setup}

Our problem formulation is a generalization of the formulation in Equation \ref{eqn:PCHA}. Instead of the archetypes learned as a linear combination of the original data points, we optimize over a general nonlinear transformation $f(X)$ from the feature space to an archetypal space in which the convex constraints are enforced. 

The generalized archetypal analysis problem is the following optimization:
{\footnotesize
\begin{equation}
\begin{aligned}
    &\underset{f,c_1,\ldots,c_k}{\argmin} 
    && \sum_{i=1}^n \| f(x_i) - \sum_{j=1}^n \alpha_{ij} c_j \|^2  \\
    & \text{subject to}
    && f\text{ is approximately invertible on }X \\
    &
    && \sum_{j=1}^k \alpha_{ij} = 1, \quad i=1,\ldots,n \\
    & 
    &&     \alpha_{ij} \geq 0, \quad i=1,\ldots,n, \; j=1,\ldots,k 
\label{eqn:gen_prob}
\end{aligned}
\end{equation}
}
The inclusion of $f$ in the optimization is unique to our formulation, while previous methods either considered no transformation (i.e., $f = \text{identity}$), or apply a fixed transformation during preprocessing (e.g. kernel PCHA). We note that our requirement that $f$ be approximately invertible is added here to allow the mapping of archetypes $\{c_j\}_{j=1}^d$ and hypothetical (convex) combinations of them to the original feature space.

\subsection{The AAnet Framework}
\begin{figure*}[!t]
  \centering
  \includegraphics[width=1\textwidth]{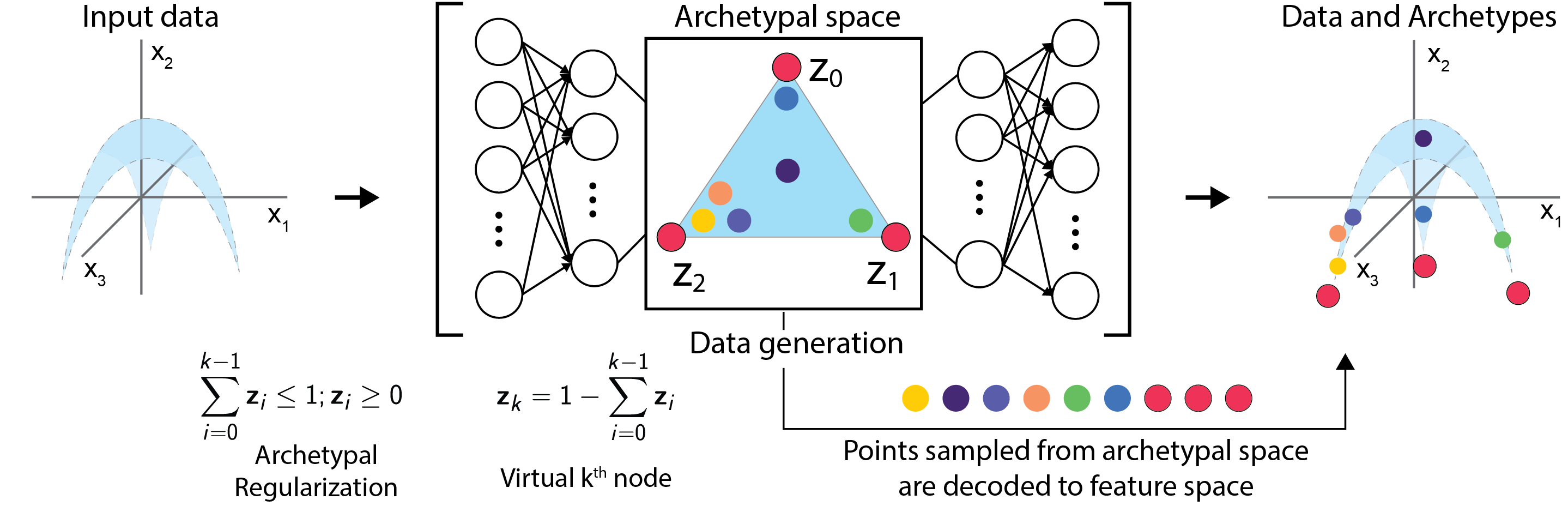}
  \caption{Illustrative representation of AAnet. AAnet learns a non-linear transformation of the input data (blue) such that within the embedding layer, the data fits well within a simplex whose vertices (red dots) represent extreme states of the data, also called archetypes. By decoding the points in the latent space, AAnet can be used for exploratory data analysis and data generation.}
  \label{fig:cartoon}
\end{figure*}

We propose a deep learning approach for solving the optimization problem in Eq.~\ref{eqn:gen_prob}, by considering $f$ as the output of a neural network we called AAnet (\textit{Archetypal Analysis network}) (see Fig. \ref{fig:cartoon}). To consider the approximate invertibility constraint, we base our network on an autoencoder, where the encoder $E(x)$ yields the transformation $f$, and the decoder $D(x)$ yields its (approximate) inverse. Then, the convex combination constraint is ensured by a novel regularization that we term \textit{archetypal regularization}. This regularization constrains the activations in that layer to be coefficients of the archetypal decomposition of a data point in the latent space of the neural network, and thus the archetypes themselves are naturally represented by one-hot vectors in this space. 

Formally, our network is formed by an encoder $z = E(x)$ and decoder $\tilde{x} = D(z)$, with the main MSE reconstruction loss: 
$
    \text{MSE} = \mathbb{E}_{x \in X}\left[\| x - \tilde{x} \|^2\right] = \mathbb{E}_{x \in X}\left[\| x = D(E(x) \|^2\right].
$
Then, to enforce $k$ archetypes, we expect $z$ to provide us with $k$ activations that sum up to one. However, notice that given such equality, we can directly compute $\alpha_k = 1 - \sum_{j=1}^{k-1} \alpha_j$. Hence, we set the embedding layer in our network to have $k-1$ nodes computed from the encoder layers, which we denote by $E^\prime(x) \in \mathbbm{R}^{k-1}$ and an additional virtual node yielding $z = E(x) = [E^\prime(x),~ 1 - \|E^\prime(x)\|_1].$ 

The described encoder architecture choice allows us to relax the unit-equality constraint to an inequality constraint, which is more suitable for the optimization used in neural network training. Therefore, our archetypal regularization is formulated as two soft constraints: 
\begin{equation}
\|E^\prime(x)\|_1 \leq 1 \text{ and } E^\prime(x_i) \geq 0,i=1,\ldots,n
\end{equation}
for every $x \in X$, which ensures the embedding layer provides convex combinations of $k$ archetypes given by the $k$ one-hot vectors of $\mathbbm{R}^k$. Note, the requirement of data points being well represented by these archetypes is implicitly enforced by the MSE reconstruction loss. The final network loss is then given by reconstruction loss $+$ two archetypal regularizations. Thus, the encoder learns a transformation that represents the data in the bounds of a convex hull, and the decode enforces accuracy of the learned representation. See Fig.~\ref{fig:cartoon} for a diagram of AAnet.

\subsubsection{Latent noise for tight archetypes}

By default, AAnet can find archetypes outside the data. However, to encourage the archetypes to be tight, \textit{i.e.} close to the data, we can add Gaussian noise $\sim N(0,\sigma)$ in the latent layer during training. Adding noise has an effect of spreading the data out in the latent space, since the autoencoder has to reconstruct points despite the noise. This, in turn, has the effect of bringing the archetypes closer to the data. We show this effect in Fig. \ref{fig:noise} where we add increasing amounts of latent noise and plot the latent archetypal space with the data and the archetypes. Finally, we note that the noise here is analogous to the $\delta$ parameter in \cite{morup2012archetypal}, which controls the distance of the archetypes to the data. By default, we set sigma such that the archetypes are close to but not significantly inside the data. In practice we set sigma such that only around 0.1 percent of the data points are outside the convex hull. For all experiments in this manuscript, this was achieved with a $\sigma$ of 0.05.


\subsubsection{Geometry based data generation}
\label{sec:methods/generate}
To generate new data using AAnet, we can sample arbitrary convex activations of the latent space and decode them to the feature space. Since this convex hull represents the boundary of the data geometry, this method allows us to sample directly from the geometry and independently of the input data distribution. For example, we can sample uniformly from the data geometry by sampling uniformly from a simplex and decode these points to the data space. Uniform sampling from a simplex was achieved by sampling from a Dirichlet distribution and then normalizing:
$
S_{ij} = \frac{-\log(U_{ij})}{\sum_{k=1}^{n_{at}} -\log(U_{ik})}, \; i = 1,\ldots,n, \; j=1,\ldots,n_{at}, 
$  
where $U$ is an $n \times n_\text{at}$ matrix whose elements are positive and i.i.d. uniformly distributed, $n$ is the number of data points and $n_\text{at}$ the number of latent archetypes. The resulting matrix $S$ is uniformly sampled on a simplex with $n_\text{at}$ corners. Finally, we get the generated data via $\hat{x} = D(S)$, where $\hat{x}$ is the generated data and $D$ is the decoder.

\subsection{Code availability}

Code and a tutorial for AAnet is publicly available on GitHub at \url{https://github.com/KrishnaswamyLab/AAnet}. This repository also includes scripts to run the quantitative comparisons included in this manuscript and to reproduce the dSprites image translation experiment.

\section{Results}
\label{sec:results}

Here we evaluate the accuracy and performance of AAnet in finding archetypes in ground-truth non-linear data with defined archetypes. We demonstrate that AAnet recovers interpretable archetypes in benchmark data from machine learning and in a biological dataset. We compare AAnet to 5 other methods. These include three linear archetypal analysis methods: \cite{morup2012archetypal} (i.e. PCHA), \cite{javadi2017non}, and \cite{chen2014fast} as well as two non-linear AA methods: kernel PCHA \cite{morup2012archetypal} and PCHA on the latent layer of a neural network \cite{wynen2018unsupervised}. For \cite{wynen2018unsupervised} we exchanged the classifier framework for an autoencoder and refer to the method as "PCHA on AE". We did this modification in order to be able to decode back to the data space, which is required for quantifying the performance of the methods, and because most of our data did not have labels. Full parameter details for AAnet are reported in Section \ref{sec:AAnet} and details of methods used for comparison are reported in Section \ref{sec:methods_comparison}.

\subsection{Archetypes from a triangle projected onto a sphere}
\label{sec:results/triangle}
\begin{figure}[tb]
  \centering
  \includegraphics[width=.49\textwidth]{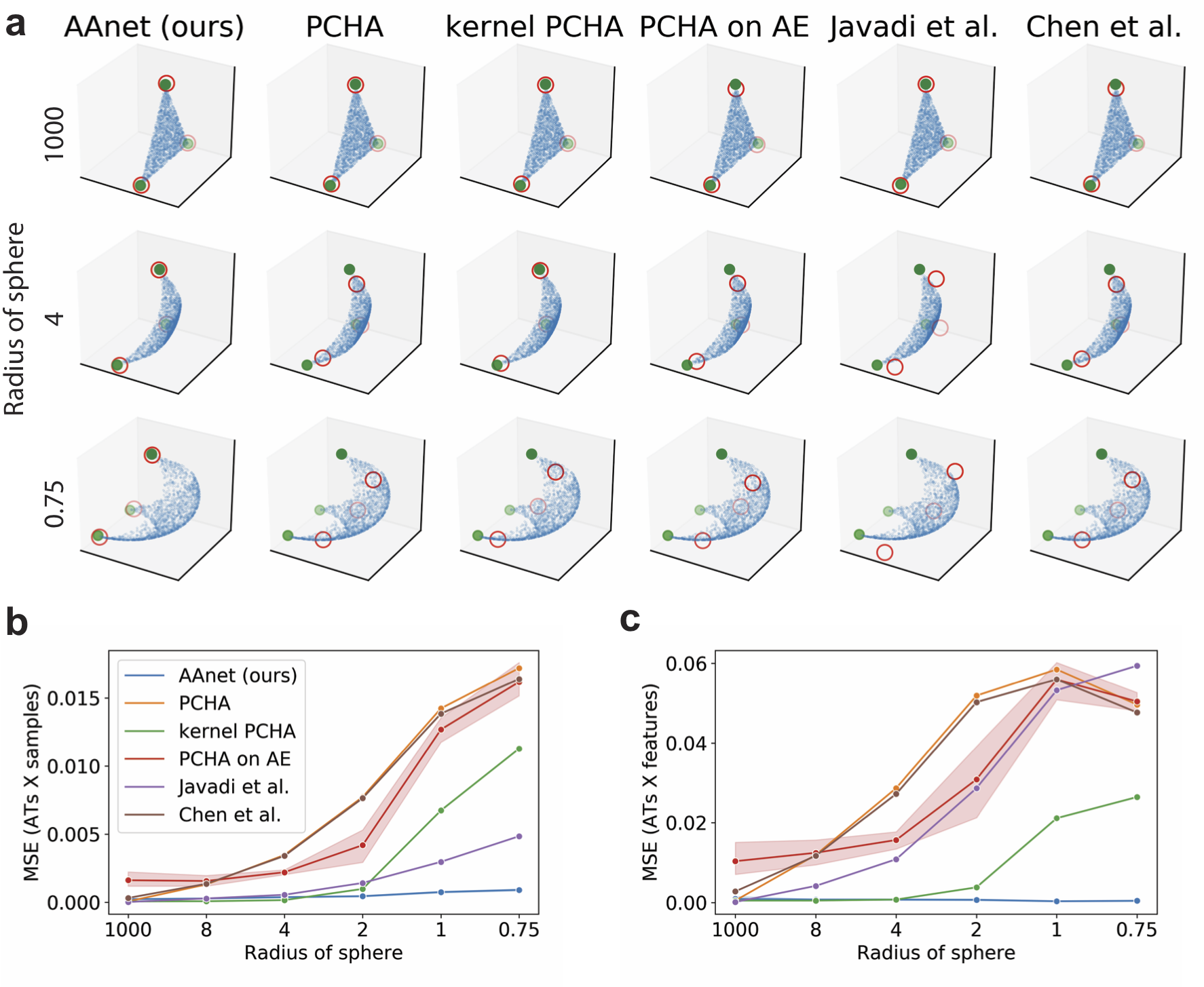}
  \caption{ (\textbf{a}) Points uniformly distributed within a triangle (blue dots) are projected onto a sphere of varying radius (columns). 6 AA methods are compared on their ability to recover the vertices of the triangle (green dots) and learn the correct mixture of archetypes for each point. Red circles mark the recovered archetypes. Right, the MSE between ground truth and recovered archetypal spaces (\textbf{b}) and recovered archetypes (\textbf{c}) are displayed for each method. Shaded area marks 95\% CI over 5 runs.}
  \label{fig:triangle}
\end{figure}

To test the ability of AAnet to find archetypes in non-linear data, we uniformly sampled 2000 points on a triangle and projected the data onto a sphere with radius $R$. To create increasing curvature on the projected triangle, we gradually decreased $R$ from 1000 to 0.75. We then ran AAnet as well as the other methods on this generated data and quantified how well each method performs by computing the MSE between the ground truth archetypes with which the data was generated and the archetypes inferred by each method (ATs x features). We also computed MSE between the recovered archetypal mixtures and the ground truth mixtures (ATs x samples). We find that with low levels of curvature all methods perform well and are able to find the correct archetypes (Fig.~\ref{fig:triangle}). However, when increasing the curvature (by decreasing the radius of the sphere) all methods other than AAnet break down, with AAnet being the only method that consistently finds the right archetypes.

\subsection{Finding archetypes of image translations}
\label{sec:img_translation}

We compared the same set of methods on the dSprites dataset, which was designed as a benchmark for disentanglement in unsupervised learning \cite{dsprites17}. The dataset consists of three image classes: rectangles, ovals, and hearts. Each class of images varies by 6 independent latent factors: horizontal and vertical offset, rotation, scale, and color. Disentanglement shares an intuitive relationship with AA, because each archetype should correspond to an extreme combination of the latent features of the dataset. Finally, although the transformations are affine in the image space, they are non-linear in the Euclidean pixel space.

\begin{figure*}[!ht]
  \centering
  \includegraphics[width=1\textwidth]{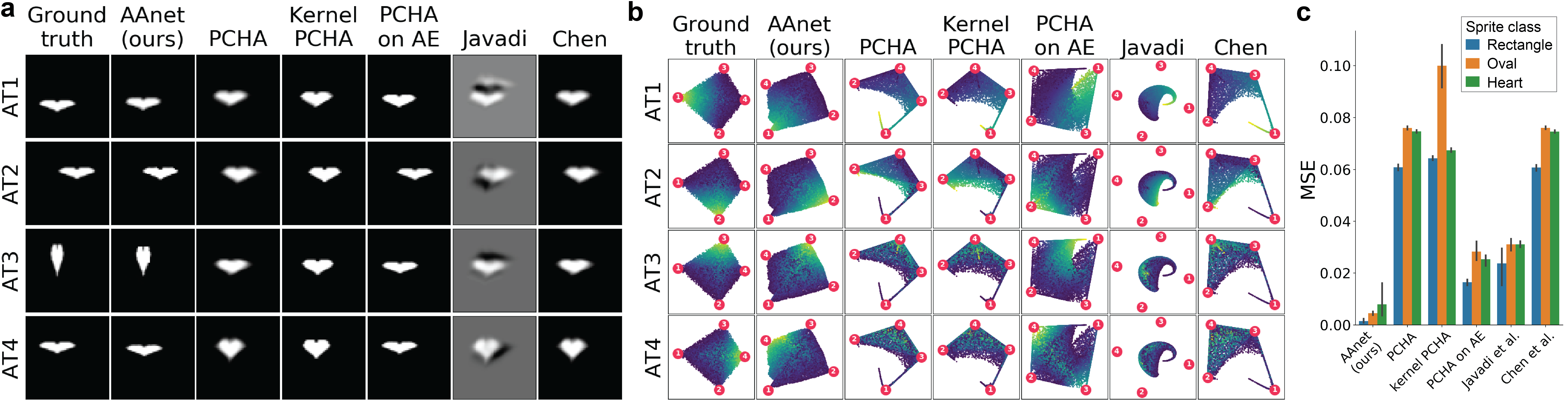}
  \caption{Comparison of AA methods on dSprites dataset. (\textbf{a}) Ground truth and recovered archetypal hearts. (\textbf{b}) Ground truth and recovered archetypal space visualized using the same test set as in \textbf{a}. Points are colored by the ground truth loading of each archetype. (\textbf{c}) Quantitative comparison of the archetypal space recovered by each method. Each method was run on 5 samples of 15,000 images using each class in the dSprites dataset. Error bars denote 95\% confidence intervals over the 5 runs.}
  \label{fig:img_translation}
\end{figure*}

To generate images for our comparison, we uniformly sampled points from a four-dimensional simplex. These values were used to adjust the horizontal offset, vertical offset, and aspect ratio for each sprite using scikit-image~\cite{waltScikitimageImageProcessing2014}. Each method was run on 5 different samples of 15,000 images for each sprite. Representative archetypes recovered from each method can be seen in Fig. \ref{fig:img_translation}a and the archetypal spaces learned for this same batch are visualized in Fig. \ref{fig:img_translation}b. A full description of the visualization algorithm can be found in Section \ref{sec:mds_plot}. To quantify the accuracy of each method, we only considered the MSE between the learned and ground truth archetypal spaces (Fig. \ref{fig:img_translation}c) because euclidean distances between images are not meaningful. We found that AAnet performed best overall, outperforming the second best method, PCHA on AE, by 80\% on average. Example images of input data and visualization of archetypes and archetypal spaces for the ovals and hearts can be found in Fig.~\ref{fig:sup_translation}.

\subsection{Generating from the data geometry with AAnet}

Next, we investigate the ability of AAnet to generate data independently of the input data density. The simplex learned by AAnet in the latent space represents the boundary of a non-linear manifold or the geometry of the data. We can sample arbitrary convex combinations of the latent space to generate data based on data geometry rather than the data density. Thus, even if the training data is non-uniformly distributed, we can learn its geometry and then sample uniformly from this geometry and decode the sample points back to the feature space.

\begin{figure}[!t]
  \centering
  \includegraphics[width=.49\textwidth]{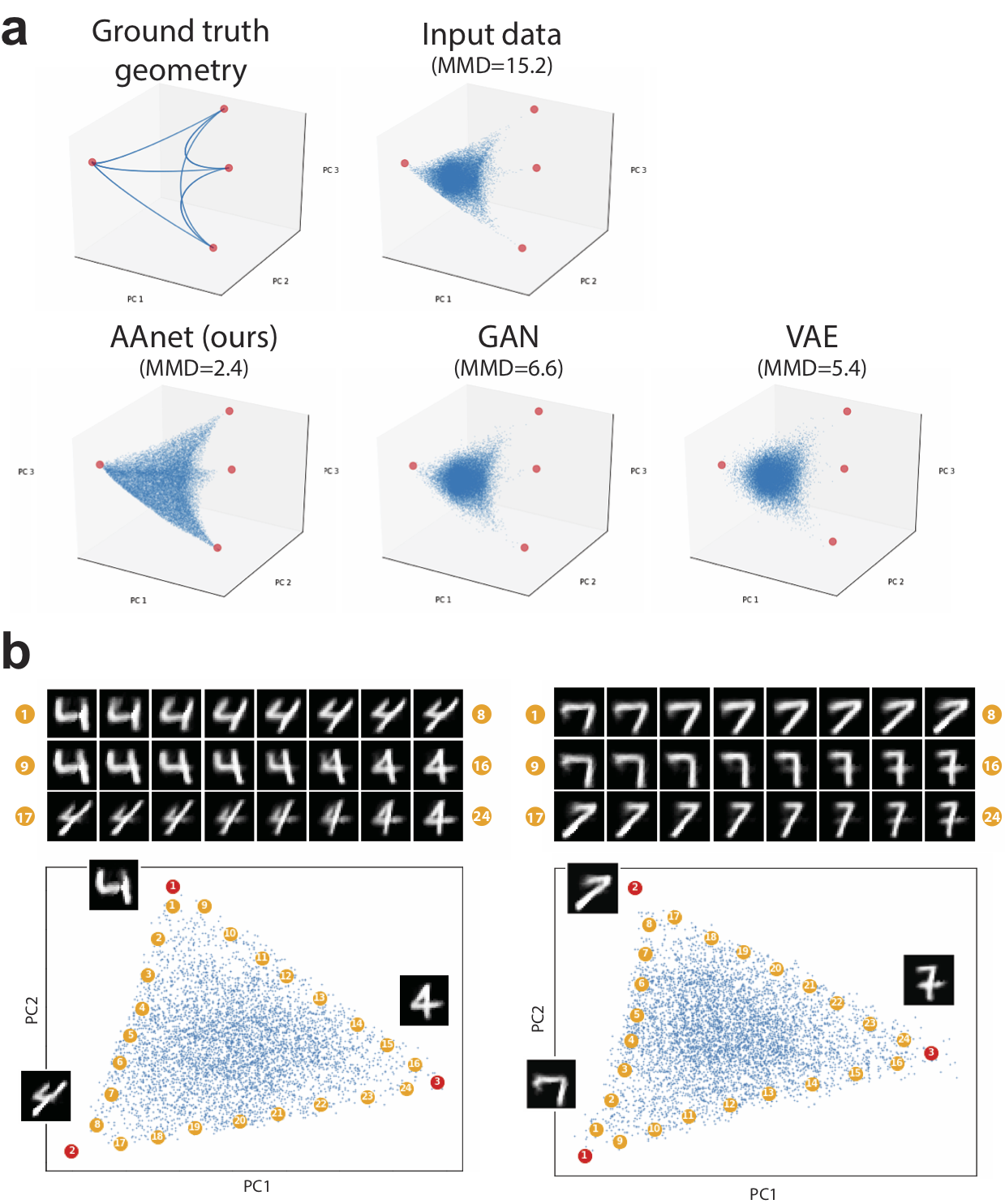}
  \caption{(\textbf{a}) Above, the ground truth data geometry and non-uniformly sampled input data. Below, data generated from AAnet, a GAN, and a VAE. MMD quantifies the discrepancy between each method and the ground truth geometry. (\textbf{b}) We uniformly sample trajectories between two archetypes from the AAnet. Shown above are such trajectories for MNIST 4s and 7s between all pairs of archetypes. Below, visualization of the archetypal space for the input data (blue), archetypes (red) with images, and sampled points (yellow).}
  \label{fig:generate}
\end{figure}

To test this, we generated a non-linear geometry with four archetypal points embedded in 100 dimensions, as shown in Fig.~\ref{fig:generate}a. We then sampled data non-uniformly (preferentially from the center) and trained AAnet, a GAN \cite{goodfellow2014generative}, and a VAE \cite{kingma2013vae} on this data. GANs and VAEs are generative models and are thus able to generate samples in the data space by sampling in their latent spaces. We then sampled from the latent spaces of these three models to generate points in the data space. The GAN and VAE both generate based on the data density, while AAnet can generate from the geometry by sampling uniformly from a simplex in its latent space (Section \ref{sec:methods/generate}). To quantify the ability of each model to generate from the geometry, we computed a Maximum Mean Discrepancy (MMD) \cite{gretton2012kernel} (using a multiscale Gaussian kernel) between the ground truth geometry and the input data, the data generated by AAnet, the data generated by the GAN, and the data generated by the VAE. AAnet had the lowest discrepancy between the generated data and the ground truth geometry performing 56\% and 64\% better than the VAE and GAN, respectively.

To demonstrate that the latent space of AAnet provides semantic structure for data generation, we sampled images by interpolating between pairs of archetypes in the latent space of AAnet trained on MNIST digits. Fig.~\ref{fig:generate}b shows this for MNIST 4s and 7s. The generated images do not appear in the training data, yet we observe gradual and meaningful transitions between them. Each interpolated image looks like a convex combination of its two corresponding archetypes.

\subsection{AAnet identifies reproducible archetypes}
\begin{figure*}[!ht]
  \centering
  \includegraphics[width=1\textwidth]{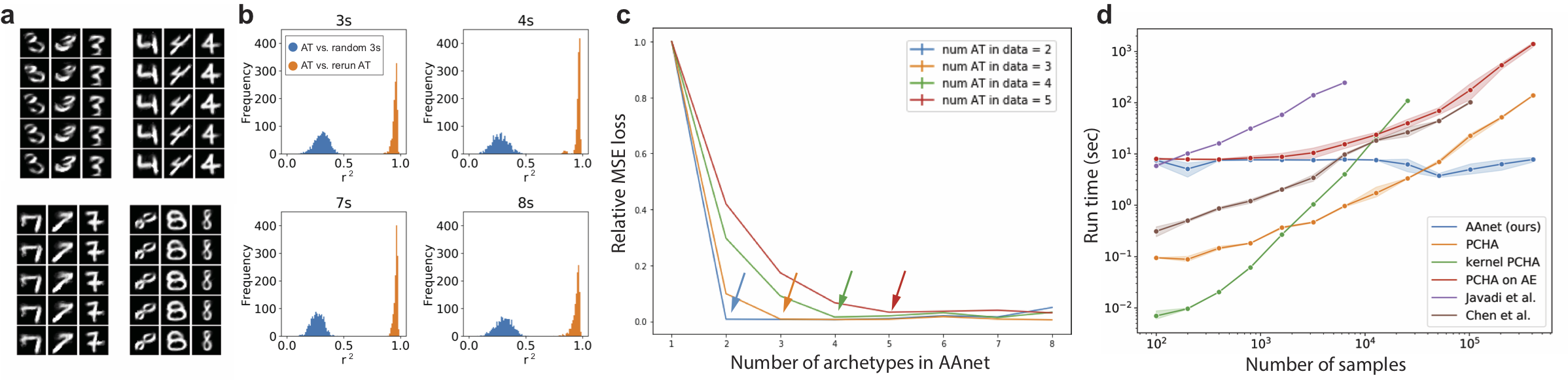}
  \caption{ (\textbf{a}) Running AAnet repeatedly on the same dataset with different random seeds identifies similar archetypes. (\textbf{b}) AAnet was run 50 times for each digit. Pearson's $r^{2}$ was calculated between all archetypes identified for each digit (orange) and random images the digit (blue). (\textbf{c}) To pick the optimal number of archetypes, we use the knee point (arrows) of loss of AAnet run on simplexes with varying numbers of vertices. \textbf{d}) Run time of AAnet and other AA methods as a function of the number of data points on generated data with 10 archetypes.}
  \label{fig:robustness}

\end{figure*}

To show that AAnet can identify robust, reproducible archetypes, we generated archetypes for each MNIST digit 50 times using different random seeds. A subset of these images are shown in Fig.~\ref{fig:robustness}a. We then calculated $r^{2}$ between archetypes identified on subsequent runs of AAnet and random MNIST images of the same digit. For all digits, we notice a significantly higher correlation between archetypes identified in subsequent runs than between archetypes and random data points (t-test, $p < 10e-16$). $R^{2}$ values are shown for a subset of digits in Fig.~\ref{fig:robustness}b. This shows that AAnet can robustly find the same set of archetypes across different runs.

\subsection{Optimal number of archetypes}

One of the main parameters in AAnet is the number of archetypes in the model. We find that the loss function of AAnet can point us to the optimal number of archetypes, \textit{i.e.} the true number of archetypes present in the data. Increasing the number of archetypes will cause the loss to decrease generally. However, the rate of decrease diminishes, with the loss converging at the right number of archetypes. To quantify this, we generated data with different numbers of archetypes (from 2 to 5) and ran AAnet with increasing numbers of archetypes in the model (1 to 8) and recorded the loss (Fig.~\ref{fig:robustness}c). We can observe an exponential decrease of the loss with increasing numbers of archetypes in the model. Indeed, the loss plateaus at exactly the correct number of archetypes which can be found using an elbow analysis. This is similar to the approach used by \cite{hart2015inferring} in which they used an elbow analysis of the explained variance by PCHA as a function of increasing numbers of model archetypes to pick the optimal number of archetypes.

\subsection{Latent noise for tight archetypes}

\begin{figure}[bt]
  \centering
  \includegraphics[width=0.5\textwidth]{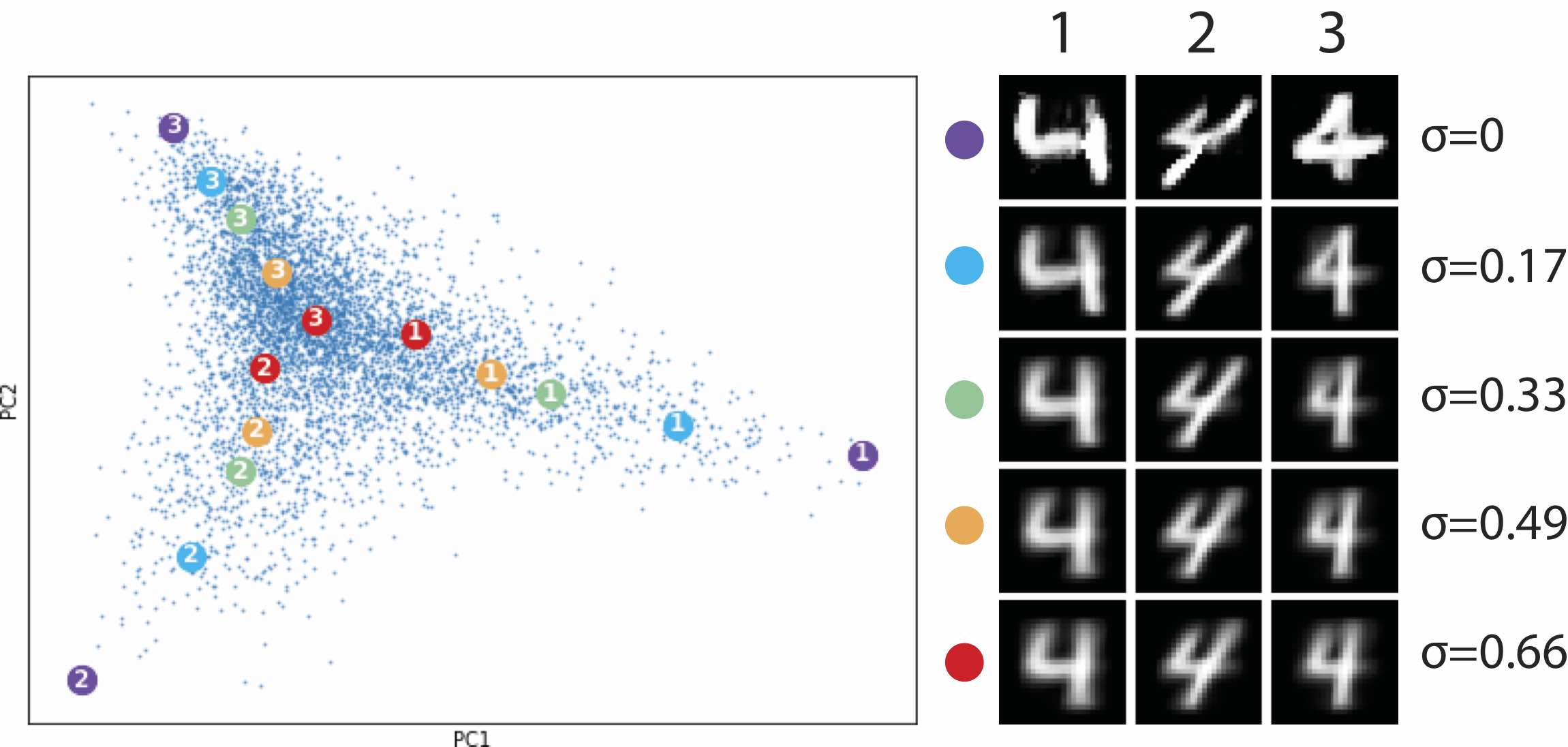}
  \caption{Adding increasing amounts of Gaussian noise (with standard deviation $\sigma$) to the latent archetypal layer causes the archetypes (circles with numbers) to come closer to (and inside) the data (blue points).}
  \label{fig:noise}
\end{figure}

Archetypes can lie far outside of the data or they can be close to data points. We are able to control the tightness of the archetypes by changing the amount of Gaussian noise we add during training to the latent archetypal layer. Increasing the noise causes the convex hull to become tighter and the archetypes to come closer to the data. To illustrate this, we ran AAnet on MNIST 4s with increasing amounts of noise (see Figure~\ref{fig:noise}). We observe that as noise increases the archetypes move closer to and inside the data. With no noise the archetypes represent hypothetical points, as they are effectively outside or in very sparse outer regions of the data. Thus, with less noise the archetypes become more extreme.

\subsection{Runtime}

Another advantage of archetypal analysis with neural networks is that it is scalable. To quantify this, we ran AAnet and the other methods on increasing sample numbers of data generated on a 10 dimensional simplex that was projected into 100 dimensions (Fig.~\ref{fig:robustness}d). While several methods run faster on smaller data (e.g. PCHA is faster or as fast up to around 50,000 samples) AAnet has the fastest run time on bigger data. In fact, the run time of AAnet is constant, while the other methods all have exponentially increasing run times with number of data points.


\subsection{Visualizing the archetypal space}
\label{sec:mds_plot}
To visualize the archetypal space, we developed a fast interpolation-based method using multidimensional scaling (MDS). First, we perform MDS on the archetypes in the feature space so that the placement of the archetypes in the plot are fixed with respect to each other. Next, the coordinates of the data in two or three dimensions are found by linearly interpolating between the coordinates of the archetypes using the archetypal mixtures learned by each method.

If $\mathbf{A}$ is the $n$-dimensional MDS coordinates of the archetypes and $\mathbf{W}$ represents that archetypal mixtures of each point in the data, then $\mathbf{X}$, the desired $n$-dimensional MDS coordinates of the data can be calculated by:

$$ \mathbf{X} = \mathbf{W}\mathbf{A}$$

\begin{figure}[htb]
  \centering
  \includegraphics[width=0.45\textwidth]{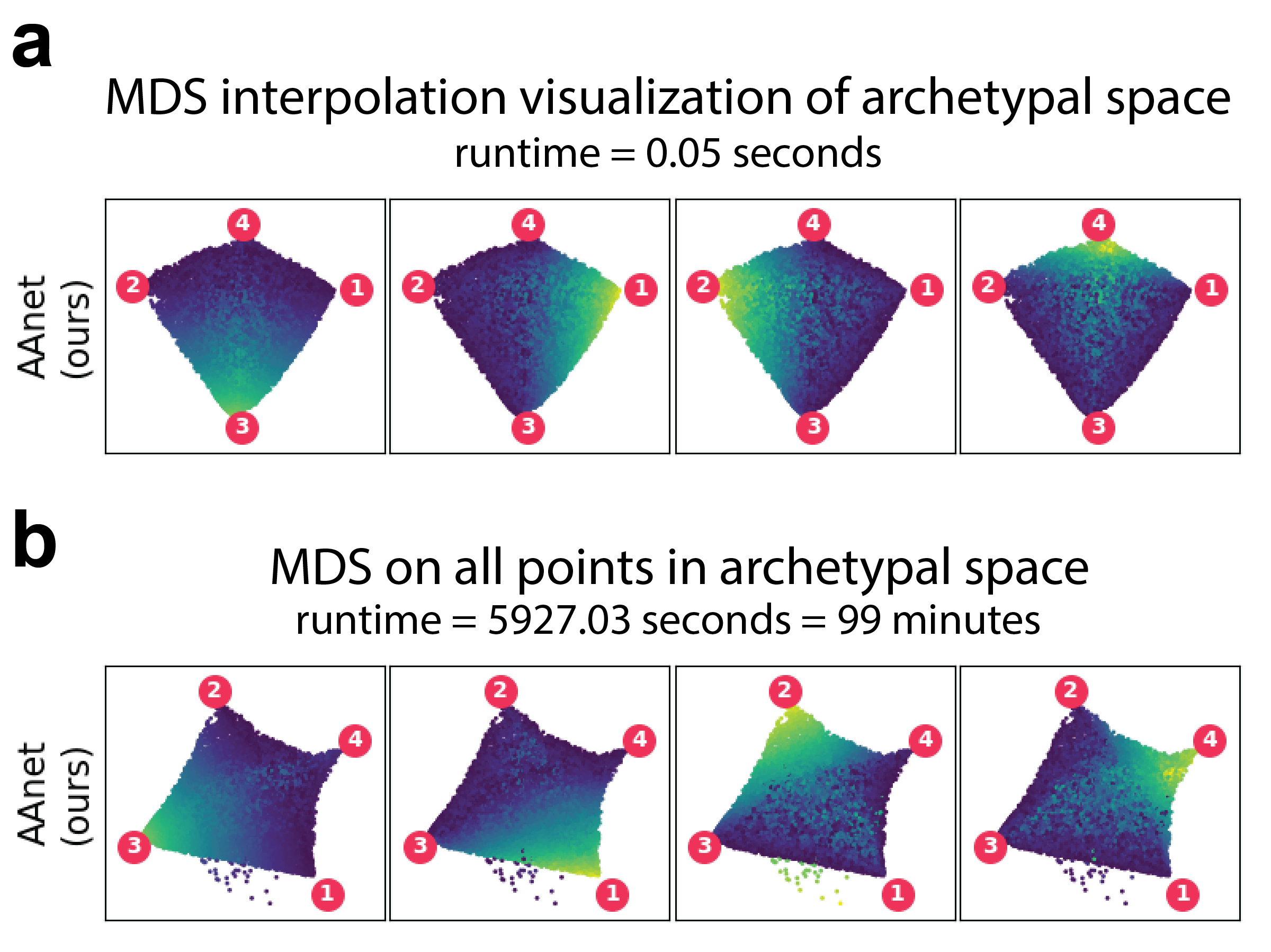}
  \caption{Comparison of our MDS interpolation method for visualizing the archetypal space to running MDS directly on all points in the archetypal space. Runtimes reflect time to calculate coordinates for 15,000 points from the dSprites experiment run on 12 cores running at 3.4GHz.}
  \label{fig:mds_viz}
\end{figure}

In practice, this interpolation method yields similar results to running MDS on a matrix comprising $\mathbf{W}$ concatenated to the archetypes along the zero-th (vertical) axis. However, this visualization method is dramatically faster. Running on 15,000 points, our method completed in 0.05 seconds to generate the coordinates show in Fig. \ref{fig:mds_viz}. Running MDS directly on all points in the archetypal space (a 15000x4 matrix), took 99 minutes to complete. We find that the results for the two visualization methods (neither of which are used for quantification) are qualitatively similar across datasets.

\subsection{Characterization of tumor-infiltrating lymphocytes using single-cell sequencing}

\label{sec:results/tcell} 
\begin{figure}[ht]
  \centering
  \includegraphics[width=.49\textwidth]{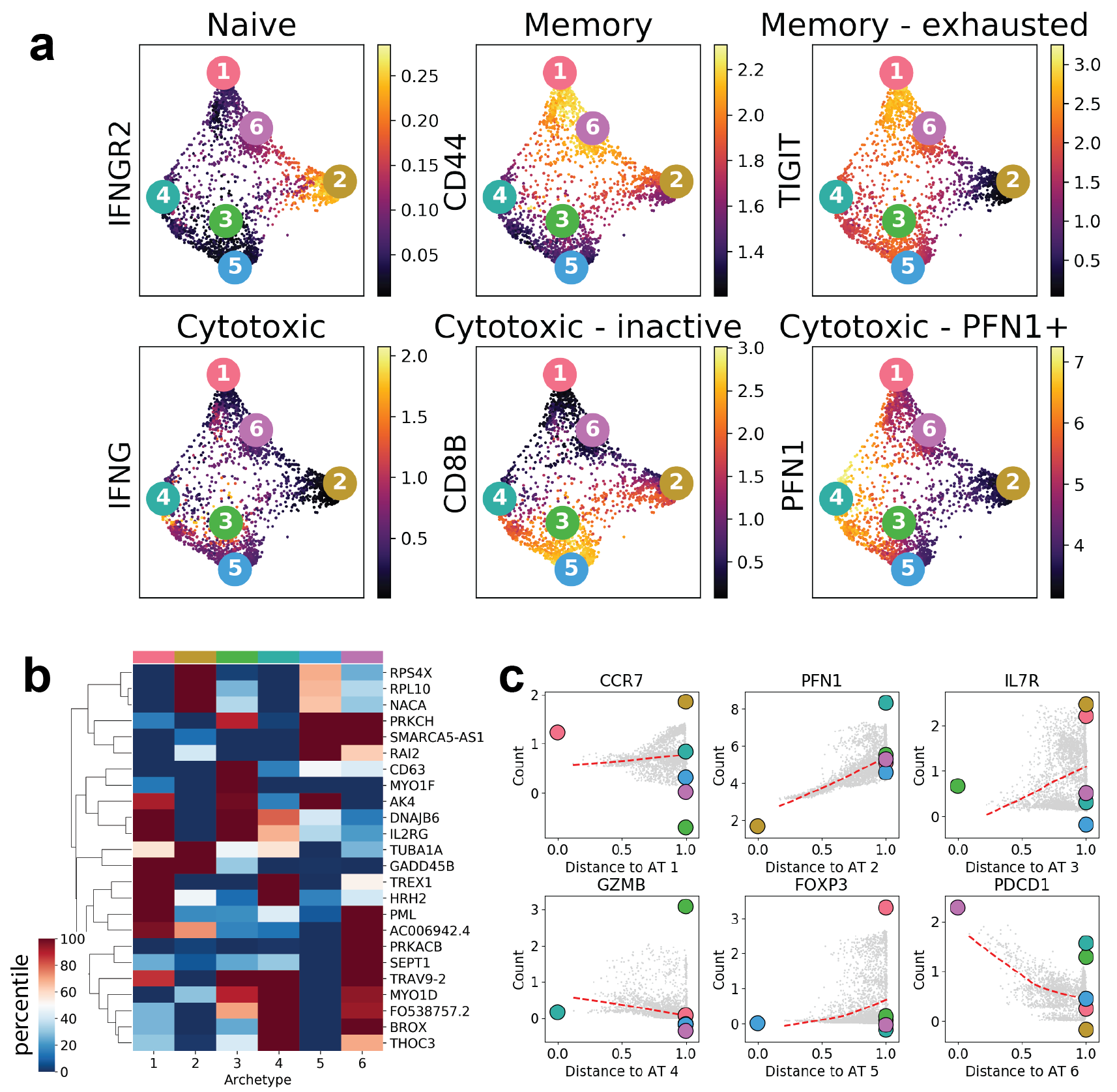}
  \caption{(\textbf{a}) PHATE visualization of scRNA-seq profiles and archetypes colored by gene expression for markers of T-cell states. (\textbf{b}) The top 5 genes from each expression signature of each archetype. (\textbf{c}) Plotting each cell (grey) by the distance to the each archetype shows how gene expression changes as distace to the archetype increases. Lowess curves (red) highlight the trends.}
  \label{fig:t_cell}
  \vspace{-7mm}

\end{figure}

Although immune cell phenotypes have classically been modelled as discrete cell states, recent applications of single-cell RNA-sequencing (scRNA-seq) have found that immune cells are better described as a continuous spectrum of states \cite{veltenHumanHaematopoieticStem2017,aziziSingleCellMapDiverse2018}. To characterize the continuous and non-linear transcriptional state space of immune cells, we applied AAnet to a newly generated scRNA-seq dataset of 3,554 lymphocytes extracted from mouse tumors and selected for expression of the T cell marker CD3. We visualized the dataset using PHATE, a dimensionality reduction method for biomedical data \cite{moonVisualizingTransitionsStructure2018}. We found that 6 archetypes best describe the dataset, with each archetype representing a specific region of the overall state space. In Fig.~\ref{fig:t_cell}a, expression of T cell marker genes is plotted on a PHATE embedding with missing gene expression values imputed using MAGIC \cite{vandijkRecoveringGeneInteractions2018}. We also found that AAnet was able to represent a relatively small subset of around 150 Cytotoxic T cells expressing interferon-gamma (IFN$\gamma$), but not profilin 1 (PFN1) (AT 3 in Fig.~\ref{fig:t_cell}a).

Next, we sought to derive a gene signature of each archetype. We decoded the archetypes into the original gene expression space and calculated the percentile expression of all genes in each archetype compared to the input dataset. Fig.~\ref{fig:t_cell}b shows the expression of the top 5 markers for each archetype. These signatures capture known markers of T cell states, such as expression of the IFN$\gamma$ receptor (IFNGR2) in archetype 2 (Naive T cells)\cite{curtsingerAutocrineIFNgPromotes2012}, high expression of perforin 1 (PRF1) in archetype 4 (Cytotoxic T cells) \cite{kagiFasPerforinPathways1994}, and upregulation of CD40L in archetype 1 (activated memory cells) \cite{makImmuneResponseBasic2006}. From these results, we conclude that AAnet is capable of characterizing the state space of a clinically-relevant biological system.

\subsection{AAnet identifies archetypal states of gut microbiomes}

\begin{figure}[ht]
  \centering
  \includegraphics[width=0.5\textwidth]{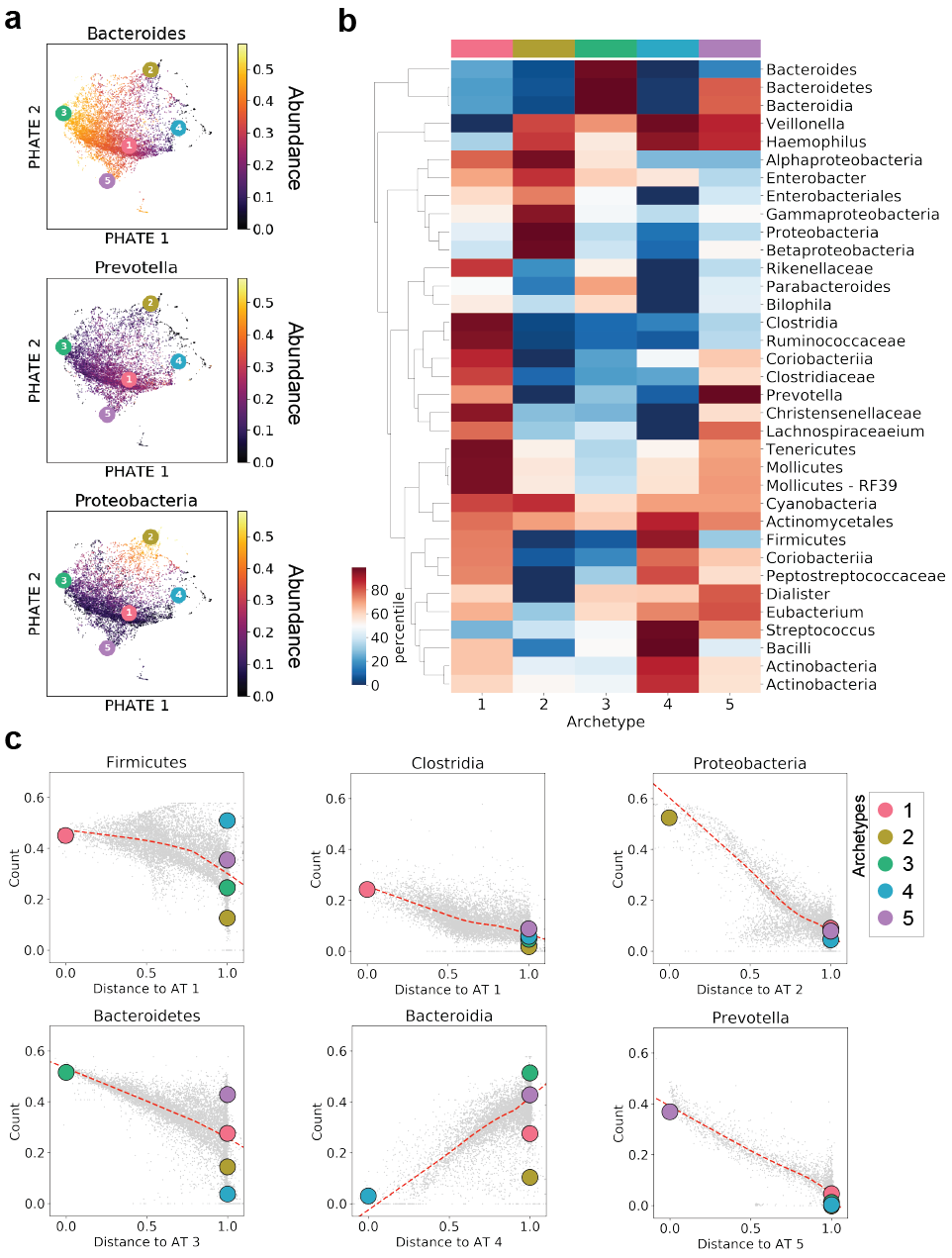}
  \caption{AAnet describes gut microbial diversity. (\textbf{a}) PHATE visualization of 8,624 gut microbiome profiles from the American Gut Project shows that AAnet captures archetypal states including the two classical Bacteriodes- and Prevotella-enriched enterotypes. (\textbf{b}) Abundance of archetypal microbial populations expressed as a percentile compared to the original data. (\textbf{c}) AAnet captures non-linear changes in microbial abundance. Here, abundance of each population within each individual (grey dots) is plotted as a function of that individual's distance to an archetype (colored dots). LOWESS on original data is plotted (red-dashed line)}
  \label{fig:microbiome}
\end{figure}

The microbiota residing in the human gut have an impact on human health, yet little is understood about the microbial diversity of the gut microbiome across individuals. Findings from the first datasets of gut microbial diversity suggested that the microbial profiles of individuals fit into one of several discrete clusters called enterotypes \cite{arumugamEnterotypesHumanGut2011}. However, more recent analysis suggests that gut diversity is better described by a spectrum of states enriched for different bacterial populations \cite{jefferyCategorizationGutMicrobiota2012, knightsRethinkingEnterotypes2014}. Recently, access to cohorts of thousands of individual microbiome profiles make it possible to understand the space of human gut microbial composition. To show the utility of AAnet in characterizing this state space, we accessed 8,624 gut microbiome profiles from the American Gut project \cite{mcdonaldAmericanGutOpen2018}. Here, bacterial diversity was determined using the 16S rRNA gene. We visualized the data using PHATE and found that the data was well described by 5 archetypes (Figure~\ref{fig:microbiome}).

Examining the abundance of various bacterial populations, we find that these archetypes represent biologically relevant microbiome states. For example, two classical enterotypes are characterized by high abundance of the Bacteroides and Prevotella genuses, respectively \cite{arumugamEnterotypesHumanGut2011}. We find that abundance of the Bacteriodes and Prevotella genuses increases in points closest to archtypes 3 and 5, respectively. This suggests that the classical enterotypes are captured by AAnet. However, we identify three other archetypes characterized by high abundance of Ruminococcaceae and Tenericutes (archetype 1), Alpha-, beta-, and Gammaproteobacteria (archetype 2), and Actinobacteria and Streptococcus (archetype 4) (Figure~\ref{fig:microbiome}b). The significance of these archetypal states remains to be investigated.

Finally, we demonstrate that the archetypes capture non-linear trends in microbial abundance. To show this, we plotted the abundance of various bacterial populations within each individual as a function of the distance of that individual to a target archetype in the latent space (Figure~\ref{fig:microbiome}c). Here, a LOWESS curve is fit to the data and plotted as a dashed red line. For example, examining abundance of the Firmicutes and Proteobacteria, we observe a clear non-linear trend in composition as individuals are increasingly distance from Archetypes 1 and 2 respectively. These results show that AAnet can be used to characterize non-linear trends across features in high-dimensional biological systems.

\section{Conclusion}
 
The main contribution of this paper is a non-linear reformulation of archetypal analysis that is solved by our neural network that we call AAnet, which features a novel archetypal regularization that enforces a convex encoding of the data in the latent layer. AAnet is an improvement over existing linear and non-linear AA methods, since AAnet 1) can learn an archetypal space even when the original data is not well fit by a simplex, 2) learns a new and optimal non-linear transformation instead of performing linear AA on a fixed non-linear transformation, such as a kernel, and 3) AAnet can generate data from a geometric description of the data \cite{lindenbaumGeometryBasedData2018} since it learns the boundary of the data geometry rather than the data density. Such descriptions are especially useful when describing biological phenotypes, since biological entities (cells, people, etc.) can exist in a non-uniform continuum of states. Using this geometric description of the data we can generate new data points by sampling uniformly from the latent archetypal space, which is useful for data that is sparse or missing in certain regions of the geometry.

\appendix
\section{Supplement}

\subsection{Neural network parameters}

\label{sec:AAnet}
The following parameters were used for experiments using AAnet. We used the same network parameters for the autoencoder networks used for PCHA on AE with two differences. First, the weights on the archetypal regularizations are set to 0 for the AE used for PCHA such that only MSE reconstruction loss was used for training. Second, we removed one hidden layer from the AE on PCHA when training on the dSprites dataset because this improved training of the vanilla AE.

For all datasets, we used 1024, 512, 256, 128 nodes in the four hidden layers of the Encoder and 128, 256, 512, 1024 nodes in the four hidden layers of the Decoder. We used between 1-8 ATs for each dataset as notes in the Results sections. All hidden layers contain LRelu activations, besides layers directly before and after archetypal layer which are linear so that each point is a linear combination of archetypes. For all but the T cell and Gut microbiome datasets, the last layer was Tanh. For the T cell and Gut microbiome datasets, a linear activation was used because these datasets were PCA reduced prior to training. The latent noise $\sigma$ was set to 0.05 for all datasets and the batch size was 256. The optimizer was ADAM, the learning rate was set to 1e-3, and the weight initialization was Xavier.

\subsection{Parameters for other methods}
\label{sec:methods_comparison}

\begin{figure*}[!t]
  \centering
  \includegraphics[width=0.93\textwidth]{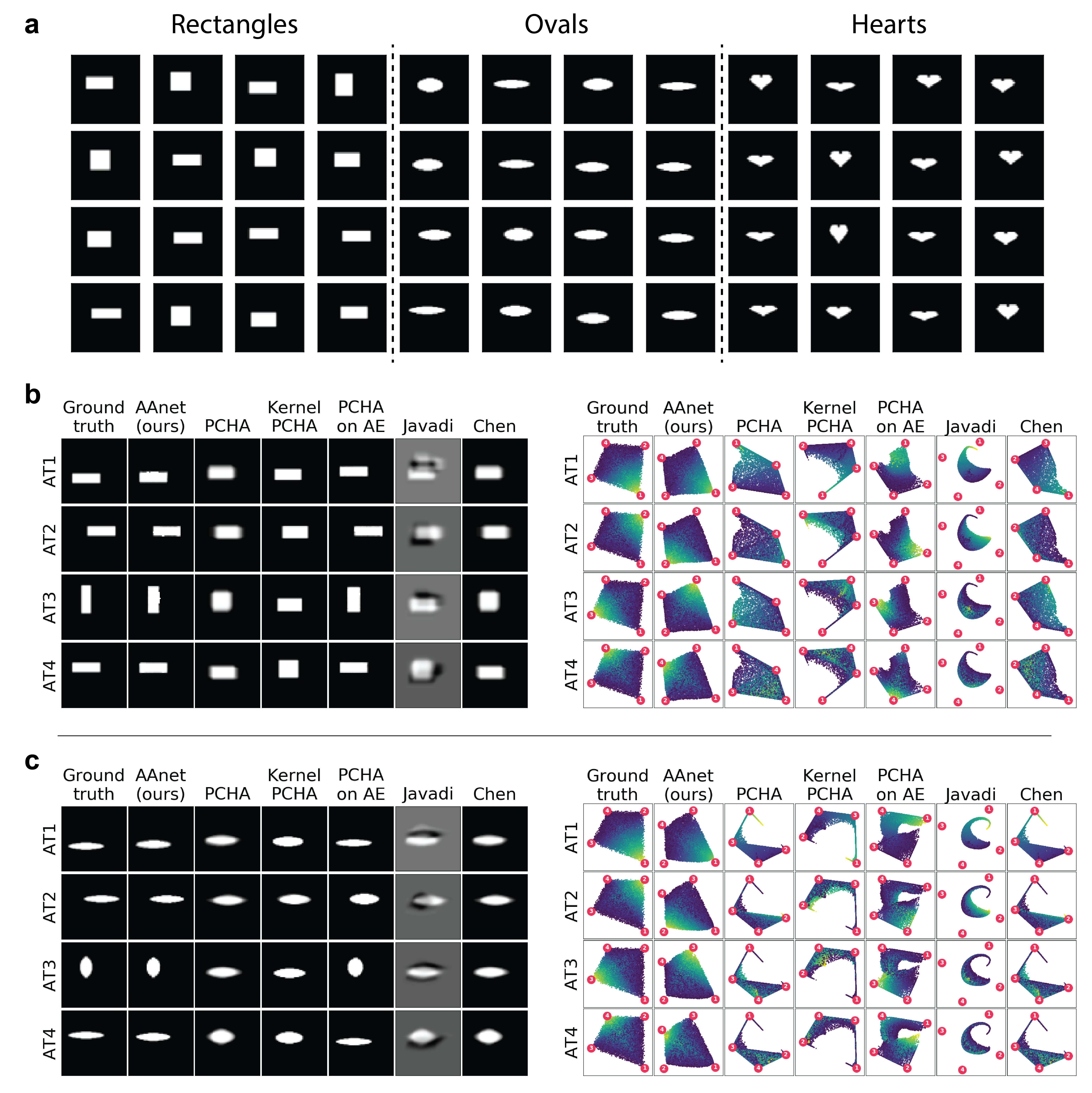}
  \caption{(\textbf{a}) Random samples of input data used for the dSprites experiment in Section~\ref{sec:img_translation}. 16 random samples of each class is shown. (\textbf{b}) Left, comparison of archetypes recovered for each method using rectangles generated with the same random seed as in Fig. \ref{fig:img_translation}). Right, visualization of the archetypal spaces (\textit{i.e.} archtypal mixtures of each point) recovered by each method.  (\textbf{c}) Same as b, but for ovals. Quantification of the accuracy of the recovered archetypal spaces can be found in Fig. \ref{fig:img_translation}c).}
  \label{fig:sup_translation}
\end{figure*}

For PCHA, we used the Python implementation of the method from \cite{morup2012archetypal} provided by Ulf Aslak and available on GitHub at \url{https://github.com/ulfaslak/py_pcha}. PCHA was run with default parameters varying only the number of archetypes as indicated in the text. To implement kernel PCHA, we first transformed the input data, $X$ with a linear kernel $XX^{\prime}$. We also tried using a radial basis kernel $exp(-((X^2)/\sigma))$ with $\sigma$ defined as the standard deviation of $X$, but this yielded exclusively higher MSE and poorer qualitative results than the linear kernel.

Implementations of the methods \cite{chen2014fast} and \cite{javadi2017non} were obtained from \url{https://github.com/samuelstjean/spams-python} and \url{http://web.stanford.edu/~hrhakim/NMF/}, respectively. Both methods were run with default parameters varying only the number of archetypes as indicated in the text.

For the GAN, we adapted code from \url{https://github.com/changwoolee/WGAN-GP-tensorflow} with a generator with dense layers: [100, 100, 100] to go from 100 dimensional Gaussian latent noise to our 100 dimensional data distribution with 4 archetypes, and discriminator with dense layers: [100, 100, 100, 1]. For the VAE we adapted code from \url{https://github.com/hwalsuklee/tensorflow-mnist-VAE} to our 100 dimensional data.

\bibliographystyle{IEEEtran}
\bibliography{main_arxiv}

\vspace{12pt}

\end{document}